\documentclass[letterpaper]{article}
\usepackage{authblk}
\usepackage{iccc}
\usepackage{graphicx}
\usepackage{amsmath,amssymb}
\usepackage{times}
\usepackage{helvet}
\usepackage{courier}
\usepackage{subcaption}
\usepackage{float}
\usepackage{lipsum}

\usepackage[colorlinks=true, citecolor=blue, urlcolor=blue, linktocpage=true]{hyperref}

\title{Dyads: Artist-Centric, AI-Generated Dance Duets}
\author[1]{\textbf{Zixuan Wang}\thanks{zwang3444@gatech.edu}}
\author[2]{\textbf{Luis Zerkowski}}
\author[3]{\textbf{Ilya Vidrin}}
\author[4]{\textbf{Mariel Pettee}}

\affil[1]{College of Computing, Georgia Institute of Technology, Atlanta, GA, 30332, USA}
\affil[2]{Faculty of Science, University of Amsterdam, Amsterdam, North Holland, 1098XH, Netherlands}
\affil[3]{Department of Theatre, Northeastern University, Boston, MA, 02115, USA}
\affil[4]{Physics Divison, Lawrence Berkeley National Laboratory, Berkeley, CA, 94720, USA}
\begin{document}
\maketitle
\begin{abstract}
\begin{quote}
Existing AI-generated dance methods primarily train on motion capture data from solo dance performances, but a critical feature of dance in nearly any genre is the interaction of two or more bodies in space. Moreover, many works at the intersection of AI and dance fail to incorporate the ideas and needs of the artists themselves into their development process, yielding models that produce far more useful insights for the AI community than for the dance community. This work addresses both needs of the field by proposing an AI method to model the complex interactions between pairs of dancers and detailing how the technical methodology can be shaped by ongoing co-creation with the artistic stakeholders who curated the movement data. Our model is a probability-and-attention-based Variational Autoencoder that generates a choreographic partner conditioned on an input dance sequence. We construct a custom loss function to enhance the smoothness and coherence of the generated choreography. Our code is open-source, and we also document strategies for other interdisciplinary research teams to facilitate collaboration and strong communication between artists and technologists.
\end{quote}
\end{abstract}

\section{Introduction}


``Dyads'' are defined as the smallest possible social group: a pair of two people. To better understand human society, it is critical to see people as not only individual actors, but also as members of relationships and larger group dynamics. Similarly, it is insufficient to view the field of dance as a discipline made up only of solo movers -- in many cultures, dance is a fundamentally social practice that relies on complex interactions between two or more people. 

Modern generative AI approaches to modeling dance, however, too often present the world of dance as a solo dancer moving in a social or even cultural vacuum. Such depictions can emerge from a combination of the technical difficulty of tracking multiple bodies at once as well as a lack of understanding of the cultural and aesthetic dimensions of dance as an art form that extends well beyond moving predictably to the beat of a pop song. 

As an embodied form that tends to favor live performance, many dances will fail to be sufficiently captured by recordings, but for artists interested in understanding their practice in part through a technological lens, there is significant potential to use technology to reimagine how we can represent and study dance in the near future.

In this paper, we present \emph{Dyads}, a generative AI model trained to model choreographic duets in contemporary dance that was built in close conversation with the dancers themselves who provided the training data. The dancers are members of the Partnering Lab, an applied research initiative focusing on embodied ethics in social interaction that is deeply interested in understanding dance partnering. In developing this model from scratch to suit the research needs of the Partnering Lab, we strove to create a research product that extended the technological boundaries of interest for our machine learning team members while centering the creative agency, data rights, and goals of the artists. 



We propose a probability-and-attention-based Variational Autoencoderspecifically designed to generate choreography for a partner based on an input dance sequence. The probability-based approach in our model addresses the deviation from the original sequence by selectively optimizing a single VAE with a certain probability. This targeted optimization ensures that each VAE independently learns more accurate and detailed representations of a dancer's movements. Meanwhile, the attention mechanism allows the model to dynamically focus on the most relevant parts of the input sequence, leading to more coordinated interactions between the two dancers. Furthermore, to improve the coherence of the generated dance sequences, we implement a custom ``velocity loss'' term in the loss function. 

Generative art has the potential to leverage formalization and abstraction to drive the creative process \cite{bisig2022generative}. Observing the current landscape of AI in art today, however, begs the question: does the integration of AI into creative processes diminish the role of human creativity, or does it augment personal artistic agency? In developing \emph{Dyads}, we not only attempt to model the communication between dance artists in duets, but we also aim to establish a collaborative workflow in which artists play a vital role  in shaping the development process of new AI methods designed to reflect their creative needs and agency. 
\footnote{\footnotesize Our open-source codebase \& demo is available at: \url{https://github.com/humanai-foundation/ChoreoAI/tree/main/ChoreoAI_Zixuan_Wang}.}

\section{Related Work}

Existing works in generative dance typically focus generating movements of a single dancer using machine learning techniques trained on 3D body joints. Crnkovic-Friis and Crnkovic-Friis \shortcite{crnkovic2016generative} proposed \textit{chor-rnn} by using Recurrent Neural Networks (RNNs) \cite{hochreiter1997long} paired with joints tracked by the Microsoft Kinect v2 sensor \cite{berger2011markerless}. The paper also presented a framework in which the artist and the model could co-create a sequence by alternating between generating sequences from each perspective.
Similarly, Pettee et. al. \shortcite{pettee2019imitationgenerativevariationalchoreography} and Papillon et. al.  \shortcite{papillon2022pirounetcreatingdanceartistcentric} developed machine learning tools based on RNNs and autoencoders and trained on motion capture data of solo improvisational dance to create authentic-looking new movements or as variations on existing ones from the real dataset. Kaspersen et al. \shortcite{kaspersen2020generative} also explored movement generation for virtual bodies with mixture density networks, autoencoders, and RNNs \cite{hochreiter1997long}.

Modeling interactions between pairs of dancers to invent new physically-plausible duet phrases presents a novel challenge in generative AI, but related tasks include more generic forms of human motion generation. Guo et al. 
 \shortcite{guo2022multi} proposed a cross-interaction attention mechanism that exploited historical information of persons who had highly correlated movements. Katircioglu et al. 
 \shortcite{katircioglu2021dyadic} introduced a motion prediction framework to model pairs of Lindy Hop dancers using a pairwise attention mechanism that allowed the model to learn the mutual dependencies in the motion history of two subjects. Li et al. 
 \shortcite{li2024interdance} also proposed a large-scale high-quality dataset called InterDance with 15 diverse genres for duet generation that capture fine-grained interactive movements.



This work presents several contributions that are distinct from prior work, including: 

\begin{itemize}
    \item An open-source model architecture that blends VAEs with Transformer decoders to model contemporary dance duets
    \item Multiple generation modes: full-duet or single-dancer conditioned on the movements of the other dancer
    \item A custom ``velocity loss'' to produce smoother generated outputs 
    \item An artist-centric model design and co-creative framework meant to reflect the specific interests of the artists.
\end{itemize}
 
\section{Methods}

In this section, we will introduce the methodologies employed to explore the integration of AI with contemporary dance duets. We detail each step of our approach from initial data extraction to the final stages of model training. The process is divided into three primary subsections: \textbf{Pose Extraction}, where we describe how movement data is processed into dynamic 3D point clouds; \textbf{Model Architecture}, which explains the design of our AI model; and \textbf{Implementation Details}, where we discuss  hyperparameter selection and optimization strategies when training our model.

\subsection{Pose Extraction}
Our dataset consists of four duet videos with around 10 minutes each ($\sim$20,000 frames) of continuous duet movements recorded at 30 frames per second. In total, the videos span 41 minutes in length. To accurately extract 3D point cloud data for each dancer, we use AlphaPose \cite{alphapose} for its robust pose estimation capabilities. AlphaPose was favored over other 2D and 3D pose estimation tools because of its superior performance in capturing dynamic 3D body movements. This tool enables precise tracking of 2D keypoints using models trained on the Halpe dataset \cite{alphapose} and 3D joints and body mesh estimation through HybrIK \cite{li2021hybrik}.

\begin{figure}[h]
    \centering
    \begin{subfigure}{\columnwidth}
    \centering
        \includegraphics[width=0.8\linewidth]{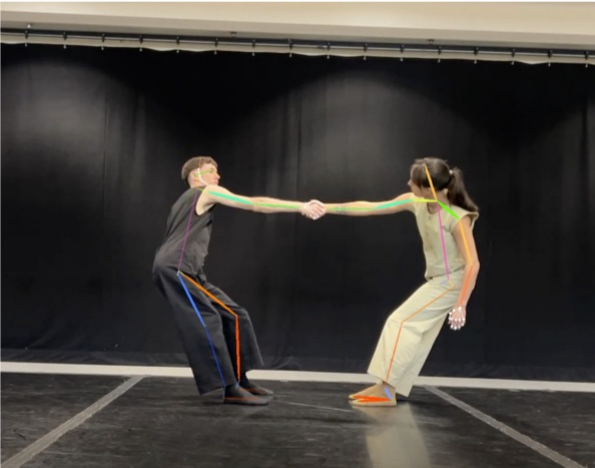}
        \caption{136 keypoints extraction result from AlphaPose.}
    \end{subfigure}

    \begin{subfigure}{\columnwidth}
    \centering
        \includegraphics[width=0.8\linewidth]{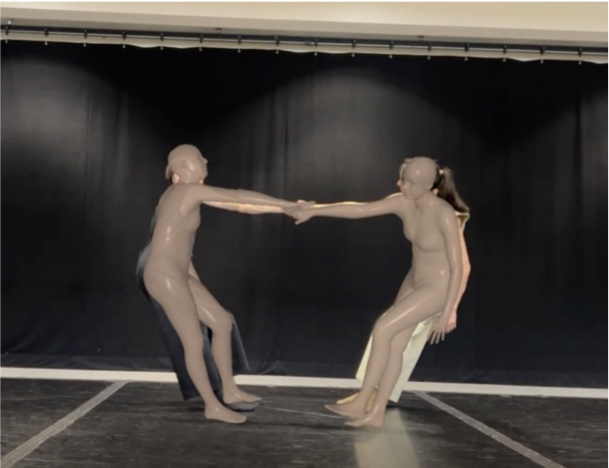}
        \caption{3D body mesh extraction result from AlphaPose.}
    \end{subfigure}
    \caption{Extraction results from AlphaPose showing keypoints and body mesh.}
    \label{fig:alphapose}
\end{figure}

We process our video data as 3D poses, each represented by 29 distinct joints as provided by HybrIK \cite{li2021hybrik} instead of body meshes due to the complexity of tracking consistent motion information along the surface of a body. 
Figure \ref{fig:alphapose} illustrates the results obtained from AlphaPose, which visualizes the 2D keypoints extraction including critical joints like the head and hands, and 3D body mesh estimation with HybrIK \cite{li2021hybrik}. We use the 3D joint data provided by AlphaPose.

In the process of converting video to 3D joints, we apply a dedicated pre-processing to address issues including:
\begin{itemize}
    \item \textbf{Missing frames:} We encounter some circumstances where no individuals are detected within certain frames. For frames in which no people are detected, we replicate the data from the previous frame to ensure continuity in the sequence.
    \item \textbf{Anomalies in person detection:} We also address another challenge in detecting the correct number of persons. In case where more than two persons are detected, we evaluate the confidence scores for each person and keep only the two highest scored detections. For frames that detect only one person, we calculate the distance between the detected person's location to two people's locations in the previous frame. To enrich the data, we replicate the position of the person farther away, ensuring that the frame reflects a more accurate representation of the scene. 
    \item \textbf{Dancers exchanging their positions:} To address this, we evaluate pairs of frames to correct potential misalignment. The algorithm computes the Euclidean distance between the position of an individual in current frame and positions of two persons in the preceding frame. The reordering is based on proximity, within the closer match being chosen as the same person.
    \item \textbf{Overall noise:} To smooth the data, we utilize the Discrete Cosine Transform (DCT). After converting to the frequency domain, all frequency components beyond the 25\% threshold are set to zero. This can effectively remove the high frequency components responsible for noise and fluctuations.
\end{itemize}

\begin{figure*}[h]
    \centering
    \includegraphics[width=\linewidth]{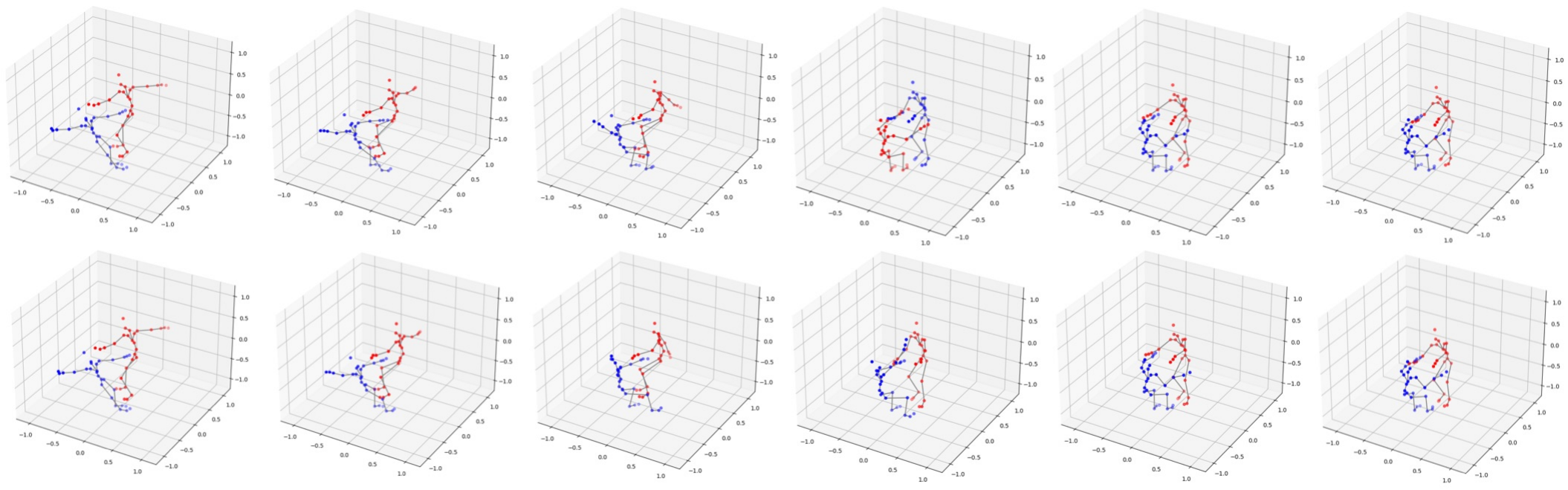}
    \caption{Movement sequence refinement illustration. The top row shows the original movement data, while the bottom row shows the refined results after processing.}
    \label{fig:processed-data}
\end{figure*}

Figure \ref{fig:processed-data} shows an example movement sequence before and after the processing stage to illustrate the improved movement dynamics. While the figure only provides a snapshot and might not fully capture all details of the enhancements, there is a distinct improvement between the original (top row) and processed data (bottom row). Particularly in the fourth image of the top row, it's evident that the two dancers have exchanged their positions, which might lead to inaccuracies in understanding the dance sequence as it affects the continuity of the movement. The processed images in the bottom row demonstrate a significant improvement in maintaining consistent dancer movements. The sequence is much more stabilized and accurate, highlighting the effectiveness of our processing techniques in handling complex spatial dynamics within dance performances.

\subsection{Model Architecture}

The generative model architecture combines elements from Variational Autoencoders (VAEs) \cite{kingma2022autoencodingvariationalbayes} with Transformer-based \cite{vaswani2023attentionneed} decoders to generate sequential dancing data for individual dancers in the duet as well as their combined interactions. The design features two VAEs to encode the movements of each dancer in the duet separately as well as a third VAE designed to encode the entire duet. Finally, Transformer decoders utilize the learned representations to predict the next time steps in an autoregressive manner. This means that each step in the sequence generation depends recursively on the previous steps, ensuring the generated sequences are logically continuous.

\begin{figure*}[htbp]
    \centering
    \includegraphics[width=\linewidth]{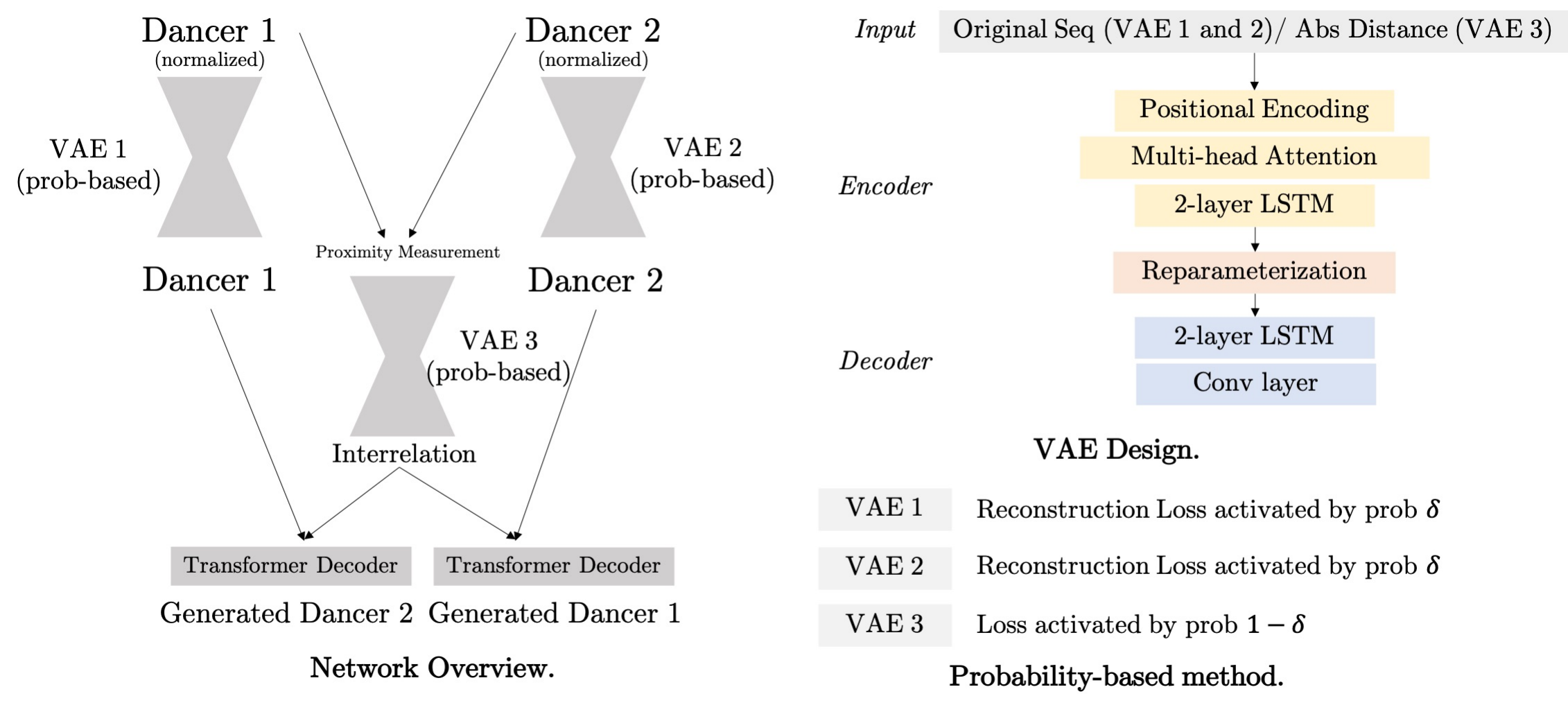}
    \caption{Network structure.}
    \label{fig:network}
\end{figure*}

Figure \ref{fig:network} shows the detailed network structure of our proposed method. The network consists of three VAEs, which are used for their capacity to serve as generative models that learn to encode data into a compressed latent space and then decode it back to the original data dimensionality. The network contains three VAEs, where VAE 1 and VAE 2 are dedicated to encoding the individual movement sequences of Dancer 1 and Dancer 2 respectively. VAE 3 focuses on the interactions and relative movements between the dancers, encoding the dynamics of the duet as a whole. It's achieved by the proximity measurement, which analyzes how closely the two dancers are by calculating the absolute distance. We believe that such a mechanism can effectively capture the nuanced interplay of coordinated dance performances. VAE 3 is a critical component that integrates the encoded information from VAE 1 and VAE 2 to maintain both the individual characteristics of each dancer and the relational dynamics in the duet.

Each VAE employs a dual objective function containing a reconstruction loss and a KL divergence loss. The reconstruction loss measures the pixel-by-pixel error between the original and reconstructed sequences, while the KL loss encourages the model to explore a diverse range of dance patterns. 

Following the VAEs, the Transformer decoders are employed to predict future dance sequences. These decoders utilize the decoded representations from VAE and apply sophisticated attention mechanisms to generate predictions. This integrated approach allows the model not only to understand the complex patterns in dance but also to innovate by generating movements that maintain the expressiveness essential to human-like dance performances.

During the experiment, we observe that supervising only the output of the Transformer decoder results in suboptimal reconstruction performance. To enhance the reconstruction accuracy of the VAEs of the individual dancers, we introduced a probability-based optimization strategy. This approach selectively focuses on the reconstruction accuracy of VAE 1 and VAE 2 under a probability $p$, effectively ensuring that the performance of each VAE is maximized. Outside of the focused training paradigm, the model will shift to a normal generation phase. This architecture is designed for maximium flexibility in generating outputs, as it has multiple modes of generation. The key generation modes are: 
\begin{itemize}
    \item \textbf{Generate a full duet:} Randomly sample the latent space of VAE 3;
    \item \textbf{Generate a partner:} Generate the movements of one dancer given the movements of the other. 
\end{itemize}

In our work, we primarily focus on the second generation mode, showing the model's ability to interpret and predict complex dyadic interactions. This mode also allows our artistic collaborators to closely evaluate the responses of the generated partners in the context of these duets.

In the following subsections, we will introduce the key components of our model, including training configuration and output generation process, the setups of the Variational Autoencoders and Transformer decoders, and the implementation of our custom loss functions.

\subsubsection{Training and Inference}
The input of the network is the full body 3D joint data for two dancers, each with size $\mathbf{X} = \{ \mathbf{S}_t, \mathbf{S}_{t+1}, ..., \mathbf{S}_{t+T} \}\in \mathbb{R}^{T\times M \times D}$, where $T$ is the length of the sequence, $M$ is the number of joints, and $D$ is the dimension of joint attributes. All data is normalized to zero mean and unit variance before fed into the model. The training of the network alternates between two modes based on a probability $p$, as we mentioned before. Specifically, under this probability, we focus on training the VAEs for the individual dancers (VAE 1 and VAE 2) with reconstruction loss and KL divergence loss. Otherwise, the dancer data $\mathbf{X}_1$ and $\mathbf{X}_2$ are fed into their respective VAEs to produce outputs $\mathbf{O}_1$ and $\mathbf{O}_2$. The interactions between the dancers represented by VAE 3 yields an output $\mathbf{O}_3$. To generate the sequence for dancer 2, the combined output from dancer 1 ($\mathbf{O}_1$) and the interaction data ($\mathbf{O}_3$) are summed to form $\mathbf{D}_1 = \mathbf{O}_1 + \mathbf{O}_3$. This data, along with dancer 2's data $\mathbf{X}_2$, is inputted into the Transformer decoder to predict dancer 2's next movement at timestep $t+T+1$. 

The target output for training, $\mathbf{Y}$, matches the input dimensionality, $\mathbf{Y} = \{ \mathbf{S}_{t+1}, \mathbf{S}_{t+2}, ..., \mathbf{S}_{t+T+1} \}\in \mathbb{R}^{T\times M \times D}$. 

During inference, particularly for generating dancer 2's sequence, the model utilizes the data of dancer 1 spanning the full sequence with length $T$, and an initial sequence of dancer 2 with length $t << T$. This provides the necessary context for the model to begin generation. Starting with the initial context of length $t$, the model predicts the next movement for dancer 2 at timestep $t+1$. This output is then recursively fed back into the model in an autoregressive manner. The model will leverage the last generated frame and append it to the preceding input data, and continue to generate subsequence movements step by step. This autoregressive mechanism allows the network to synthesize dance sequences of arbitrary lengths up to $T$, enabling the model to continuously produce movements without predefined constraints on sequence duration. This capability provides a robust framework for those applications where the sequence duration may vary.

\subsubsection{Variational Autoencoder Configuration}
Our network has three specialized VAEs, each designed to process different aspects of dance data. These VAEs share a common architecture, including an encoder, a decoder, and advanced neural network techniques to support the complex task of movement generation.



Each encoder begins with a positional encoding, which uses sine and cosine functions of different frequencies to embed temporal information into the input sequence \cite{vaswani2023attentionneed}:
\begin{align}
    PE(\text{pos}, 2i) &= \text{sin}(\text{pos} / 10000^{2i/d_{\text{model}}}) \\
    PE(\text{pos}, 2i+1) &= \text{cos}(\text{pos} / 10000^{2i/d_{\text{model}}})
\end{align}

This encoding provides the model with crucial information about the sequence order of the dance movements, which is essential for understanding the temporal dynamics of the dance. Following positional encoding, we use a multi-head Attention mechanism that enables the model to focus on different parts of the dance sequence simultaneously, improving the ability to model complex dependencies.

A two-layer Long Short-Term (LSTM) network complements the attention mechanism, helping to capture temporal dependencies over time. LSTMs are particularly suitable for processing sequences where past information is essential for future prediction.

Central to each VAE is the reparameterization trick, which facilitates the generation of new dance sequences. This technique allows for the efficient sampling of latent variables from the learned distribution, which are then used by the decoder to reconstruct or generate new dance sequences. This step is critical for ensuring variability and creativity in the generated dance movements, as it enables the model to explore a wide range of possible dance sequences within the learned latent space. 

In detail, the encoder outputs the mean $\mu$ and variance $\sigma^2$ of a Gaussian distribution over latent variables. To generate a sample $z$ from the latent space that is differentiable with respect to the network parameters, we apply the reparameterization trick, where $z$ is expressed as 
\begin{equation}
    z = \mu + \sigma \cdot \epsilon
\end{equation}
where $\epsilon$ is a noise variable sampled from a standard normal distribution $\epsilon \sim \mathcal{N}(0, I)$. This reparameterization approach allows our model to perform gradient-based optimization effectively, thus enabling the generation of diverse and innovative dance sequences that are crucial for creative choreography applications.

The decoder consists of a two-layer LSTM to maintain the temporal coherence in the reconstructed sequences. A convolutional layer follows the LSTM output, which serves to smooth and refine the decoder outputs. This convolutional layer helps reduce anomalies in movements, ensuring the generated dance sequences are realistic.



\subsubsection{Transformer Decoder}
We use a standard PyTorch \cite{paszke2019pytorchimperativestylehighperformance} Transformer decoder architecture to generate data based on both the original sequences and the context provided by three VAEs. 

In our configuration, the Transformer decoder integrates two key inputs: the \texttt{target} and the \texttt{memory}. The \texttt{target} consists of the original dance sequence data for the dancer whose movements are being predicted. The \texttt{memory} input is the aggregated output from the individual dancer's VAE and VAE 3. 

Initially, both \texttt{target} and \texttt{memory} are fed into a linear transformation to ensure they conform to the dimensionality requirements of the subsequent transformer layers. Positional encoding is then added to the \texttt{target} sequence, embedding temporal context into the data.

Finally, the Transformer decoder processes the \texttt{target} and \texttt{memory} through its multi-head attention and feed-forward layers to generate the movement in the next time step. The output will pass through another linear layer designed to ensure smoothness.

\subsubsection{Losses}
For VAE, we use MSE loss and KL divergence loss. MSE loss is used to measure the average squared difference between the predicted outputs and the actual target values, which is defined as $l_{mse}=mean(||\mathbf{\hat{Y}}_t - \mathbf{Y}_t||_2^2)$. Here, $\mathbf{\hat{Y}}_t$ is the output of the network at time $t$. 

KL divergence loss is crucial for the training of VAE, which measures the difference between the learned distribution and the prior distribution, typically assumed to be a standard normal distribution:
\begin{align*}
    l_{kl} &= \text{KL Divergence}(\mathcal{N}(\mu, \sigma^2)|| \mathcal{N}(0, I))  \\
    &= \frac{1}{2} \sum_{i=1}^{d} (\mu_i^2 + \sigma_i^2 - \log(\sigma_i^2) - 1)
\end{align*}
where $d$ is the dimension of the latent variable, with $\mu$ and $\sigma$ representing the mean and standard deviation vectors produced by the VAE encoder, respectively.

Despite these losses, the model still suffers from jitter. To resolve this, we introduce a velocity loss, which aims to maintain the continuity of motion between consecutive frames. It calculates the first-order difference between frames, referred to as velocity $\mathbf{v}_t =\mathbf{\hat{Y}}_t - \mathbf{\hat{Y}}_{t-1}$, and measures the change in this velocity over a specified number of frames (denoted by \texttt{frames})


\begin{align}
    l_{velocity} = mean(||\Delta\mathbf{v}_t||_2)
\end{align}
where $\mathbf{v}_t=\mathbf{x}_{t+1} - \mathbf{x}_{t}$, and $\Delta\mathbf{v}_t=\mathbf{v}_{t+frames} - \mathbf{v}_t$.

The total loss is defined as
\begin{align}
    l = \alpha * l_{mse} + \beta * l_{velocity} + \eta * l_{kl}
\end{align}
where $\alpha, \beta, \eta$ are hyperparameters that balance the influence of each loss component on the training process. 

\subsection{Implementation Details}

We selected hyperparameters using a grid search. The resulting hyperparameters were $T=64$, number of frames $=1, p=0.1, \alpha=0.5, \beta=0.05, \eta=0.00005$. 

During the experiments, test results show that the model tends to generate similar movement patterns across diverse test datasets. To address this issue, we implement data augmentation techniques, specifically introducing Gaussian noise into the training data to enhance the diversity of the training samples and encourage the model to learn a broader range of movement dynamics. The standard deviation for the Gaussian noise to implement data augmentation is 0.01. 

The number of heads for multi-head attention is 8 and the dimension of embedded feature fed into MHA is 64. The latent dim for VAE is also 64.

The models were trained on 1 NVIDIA A100 GPU over a span of 13 hours within 100 epochs. We use Adam optimizer with $\beta_1,\beta_2=(0.9, 0.999)$ and $lr=0.001$. Furthermore, we integrate a Cosine Annealing scheduler for the learning rate with $T_{\text{max}}=100$. 

\section{Results}


In this section, we present the results of our experiments both quantitatively and qualitatively. 

\subsection{Quantitative results}

To quantitatively evaluate our predicted results, we calculate the Mean Squared Error (MSE) between the predicted sequences and actual movements for each dancer in the duet. We track the MSE at multiple points along sequences of length $T=64$ to examine how errors accumulate over time. Specifically, we select 10 random  sequences from our test data and calculate the average MSE at four time steps: $t=16, t=32, t=48,$ and $t=64$. The results are shown in Table \ref{tab:mse_values}.

\begin{table}[ht]
\centering
\caption{Mean Squared Error at various sequence lengths.}
\label{tab:mse_values}
\begin{tabular}{cc}
\hline
\textbf{Sequence Length (t)} & \textbf{MSE} \\
\hline
16 & 0.0126 \\
32 & 0.0197 \\
48 & 0.0219 \\
64 & 0.0263 \\
\hline
\end{tabular}
\end{table}

At $t=16$, the MSE was relatively low, suggesting that the model performs more accurately in the short term. As time progresses, the MSE gradually increases, indicating a trend in which the predictions increasingly differ from the true data. This pattern suggests that while the model is capable of reproducing the dynamics of dance movements effectively in the initial phases of the sequence, it increasingly struggles to retain this accuracy for predictions further in the future . It also indicates potential areas for improvement in future model development, possibly through more sophisticated temporal modeling architectures.

\subsection{Qualitative results}
\begin{figure*}[h!]
    \centering
    \includegraphics[width=\linewidth]{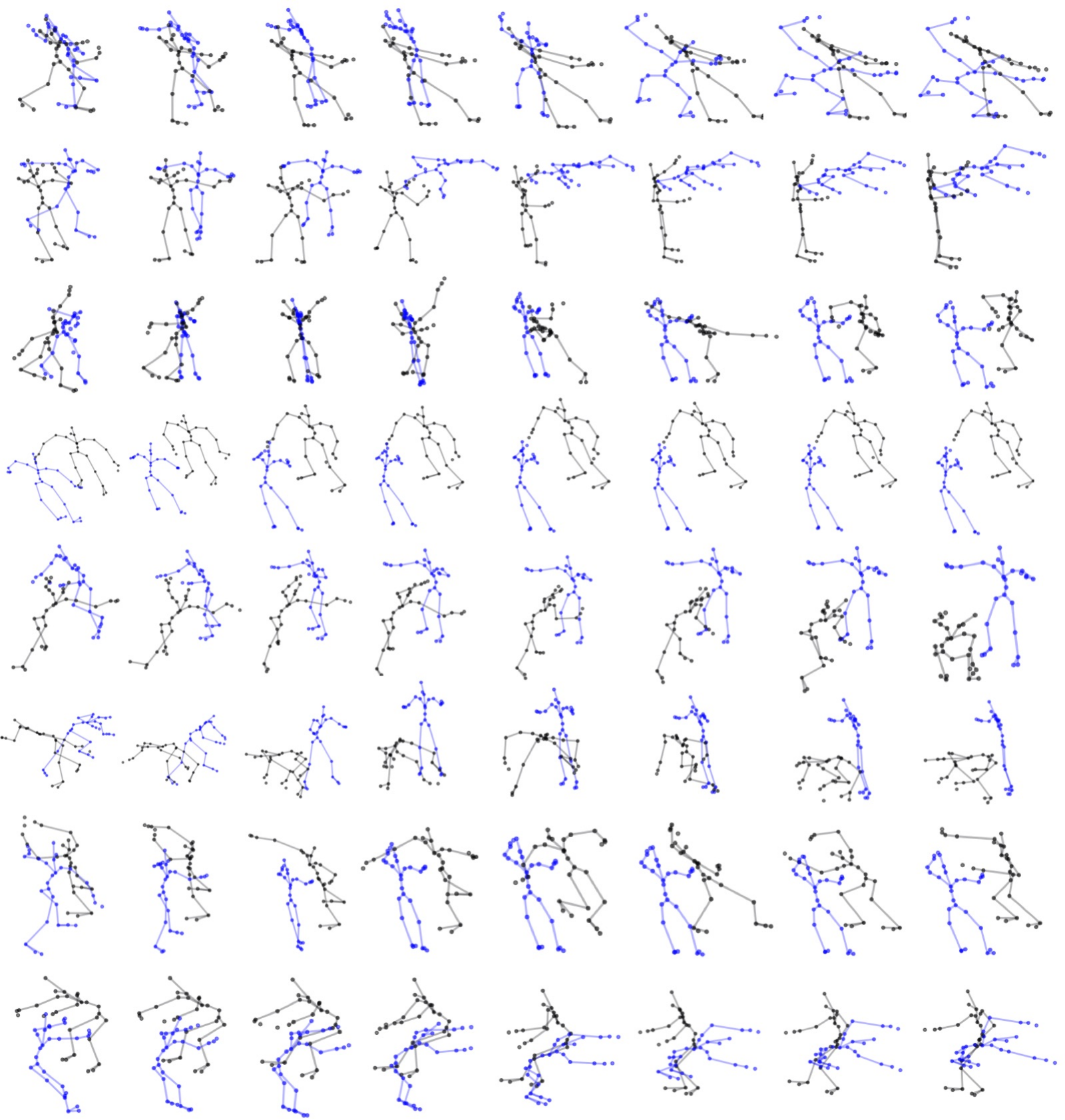}
    \caption{Examples of several generated test sequences from the model in which the figure in blue is a generated partner conditioned on the movements of the figure in black. The movements range in physical plausibility, ranging from slow and deliberate movements to more complex and even, at times, unphysical dynamics.}
    \label{fig:result}
\end{figure*}

Figure \ref{fig:result} shows a number of the generated results from our model for which the dancer in blue is generated, conditioned on the real input sequence of the dancer shown in black. Each row represents a single movement sequence.

The generated movements are diverse, reflecting the variety of movements of the other dancer in each duet. The generated dancers tend to face or otherwise mirror the other dancer. Many generated movements are relatively slow or subtle and focus on e.g. deepening a bend in a pli\'{e}, as shown in the sequence in the final row of Figure~\ref{fig:result}. Some movements show rotations in space, kicking movements with the feet, and stretching of outreached arms. In some cases, however, it is clear that the generated movements do not match the speed or complexity of the partner's movements, indicating that the model cannot always effectively model more unusual movement dynamics (see e.g. the penultimate row of Figure~\ref{fig:result}). More dramatically, in the second row, the model shows a floating effect where the generated dancer appears to have no contact with the ground. This phenomenon is not physically plausible and suggests challenges in imposing gravitational constraints on the model. In the third row, the sequences display excessive rotations that surpass the natural range of human movement. 

For the collaborating artists whose movements were captured in these videos, these generated outputs often reflected some of the principles of their practice, including a reciprocal focused attention on one another to avoid injury and embody  a sense of mutual care. On the other hand, the wide variability in the generated outputs was seen as both surprising and generative. The most unconventional movements prompted exciting discussions about what partnering should look like, how much the two movements should physically mirror one another, and to what extent gravity, anatomy, and friction should dictate what ``good'' partnering looks like.

\vspace{0.1cm}

\section{Discussion}

\vspace{0.1cm}

\subsection{Artist Involvement}

The relationship with our artistic colleagues was central to our development of this model, as we believe all creative AI work should directly engage with practicing artists. Using an artist-centric framing for this project meant that the artists not only gave explicit consent to process their data, but also helped in defining the scope of data collection and usage, designing the settings for data collection, defining core vocabulary terms for both the AI and dance researchers, iterating on the representation of the extracted 3D poses, model development and refinement, and discussion of the results. Another critical framing for this artist-technologist collaboration was recognizing each field as equally rigorous and important investigative frameworks. We approached the chosen task not as a ``problem'' for AI or technology to ``solve'', but as an opportunity to ask new questions. \\

\pagebreak

\clearpage

\subsection{Future Directions}
\begin{itemize}
    \item \textbf{Model Expressiveness:} Enhancing the expressiveness of the model is essential for generating more complex and dynamic sequences. This could involve improving the model's ability to capture complex correlations between the two dancers.
    \item \textbf{More Data Collection / Augmentation:} Since the amount of data is limited in our project, we could continue to source new input data from our artistic partners to improve the overall results and/or implement advanced data augmentation strategies such as geometric transformations and noise injection to improve the model's ability to capture more varied movements.
    \item \textbf{Real-time Choreography Generation:} Developing a system for real-time dance choreography generation would open up thrilling new dimensions of live performance by e.g. producing a generated partner for a solo dancer. This would require reducing overall latency in model inference and configuring a real-time input data stream using pose estimation software.
    \item \textbf{Group Interactions:} Developing a user-friendly interface that allows choreographers and dancers to easily train their own duet models and interact with the outputs would provide new avenues for artists to imagine new movements of not only duets, but also group choreographies. 
\end{itemize}

\section{Conclusion}
In this paper, we propose a probability-and-attention-based Variational Autoencoder for generating a choreographic partner conditioned on an input dance sequence. Our model design was closely shaped by the goals of our artistic collaborators, who are interested in understanding the ethics of partnering in dance. While the results show some significant variation, in general they demonstrate that modeling contemporary dance duets with Transformer-based architectures can effectively produce conditional dance partner movements that are not only physically plausible, but also reflect core tenets of the dance training of the original artists. We emphasize the importance of developing creative AI models in close collaboration with artists in order to effectively center creative research questions over purely technological ones within an overall ethical framework.

\bibliographystyle{iccc}
\bibliography{iccc}

\end{document}